\documentclass[journal]{IEEEtran}

\ifCLASSINFOpdf
\else
   \usepackage[dvips]{graphicx}
\fi
\usepackage{cite}
\usepackage{amsmath,amssymb,amsfonts}
\usepackage{algorithmic}
\usepackage{graphicx}
\usepackage{textcomp}
\usepackage{xcolor}
\usepackage[hidelinks]{hyperref}

\hypersetup{
    colorlinks=true,   
    linktoc=all,        
    linkcolor=blue,     
}
\ifCLASSINFOpdf
\else
\fi

\usepackage{amssymb}
\usepackage{multirow}

\usepackage{url}

\usepackage{graphicx}
\usepackage{booktabs}
\usepackage{color}
\usepackage[justification=centering,font=footnotesize,labelsep=period]{caption}
\begin{document}
%

\title{KAN See In the Dark}


\author{\IEEEauthorblockN{Aoxiang Ning,
Minglong Xue\IEEEauthorrefmark{1}, Jinhong He and Chengyun Song}

\thanks{
This work is supported by the Natural Science Foundation of China under Grant (62472059), the Science and Technology Research Program of Chongqing Municipal Education Commission (KJQN202401106), the Special Project for the Central Government to Guide Local Science and Technology Development (2024ZYD0334), the Chongqing University of Technology High-quality Development Action Plan for Graduate Education (gzlcx20243151).\par (Corresponding author: Minglong Xue)
Aoxiang Ning, Minglong Xue, Jinhong He and Chengyun Song are with Chongqing University of Technology, Chongqing, 400054, China. (e-mail: ningax@stu.cqut.edu, xueml@cqut.edu.cn, hejh@stu.cqut.edu.cn, scyer123@cqut.edu.cn)

}
}


%



\maketitle

\begin{abstract}
Existing low-light image enhancement methods are difficult to fit the complex nonlinear relationship between normal and low-light images due to uneven illumination and noise effects. The recently proposed Kolmogorov-Arnold networks (KANs) feature spline-based convolutional layers and learnable activation functions, which can effectively capture nonlinear dependencies. In this paper, we design a KAN-Block based on KANs and innovatively apply it to low-light image enhancement. This method effectively alleviates the limitations of current methods constrained by linear network structures and lack of interpretability, further demonstrating the potential of KANs in low-level vision tasks. Given the poor perception of current low-light image enhancement methods and the stochastic nature of the inverse diffusion process, we further introduce frequency-domain perception for visually oriented enhancement. Extensive experiments demonstrate the competitive performance of our method on benchmark datasets. The code will be available at: \href{https://github.com/AXNing/KSID}{https://github.com/AXNing/KSID}.
\end{abstract}



\IEEEdisplaynontitleabstractindextext

%
\IEEEpeerreviewmaketitle

\section{Introduction}
Low-light image enhancement (LLIE) is a critical task in computer vision and is essential for various applications ranging from surveillance to autonomous driving\cite{li2021low}. Images captured in low-light environments often suffer from low contrast and loss of detail, making downstream tasks such as object or text detection, semantic segmentation, and others highly challenging\cite{xu2024seeing}. Therefore, to further enhance various visual applications in poor environments, low-light image enhancement tasks have received extensive attention from researchers\cite{he2024zero,xue2024low,singh2024illumination}.\par
Traditional methods utilize retinex theory\cite{guo2016lime} and gamma correction\cite{rahman2016adaptive} to correct image illumination. With the development of deep learning, some methods\cite{zhang2023multi,zhang2023lrt,xu2023low,cai2023retinexformer} significantly improve the performance of low-light image enhancement by learning the mapping between low-light and normal images in a data-driven way. Recently, the diffusion model \cite{ho2020denoising,nichol2021improved} has received much attention for its remarkable performance in generative tasks. \cite{fei2023generative} introduced the diffusion model into the low-light image enhancement task to improve the recovery of image details and textures in low-light conditions. \cite{10663248} leverages the generative capabilities of the latent diffusion model to accelerate inference speed while achieving excellent perceptual fidelity. \cite{hou2024global} establishes a global structure-aware regularization that promotes the retention of complex details and enhances contrast during the diffusion process, further improving image quality. \cite{xu2024upt} proposed a multi-scale Transformer conditional normalized flow (UPT-Flow) based on non-equilibrium point guidance for low-light image enhancement. However, there are nonlinear degradation factors in low-light enhancement tasks, such as uneven illumination and varying degrees of noise in low-light images, and it is difficult for existing methods to model complex nonlinear relationships on limited data. On the other hand, although current state-of-the-art technologies have achieved remarkable breakthroughs in the field of LLIE, its inner workings are often viewed as a black box that is difficult to decipher, limiting the development of the model in specific domains. \par
The recent introduction of Kolmogorov-Arnold Networks\cite{liu2024kan} has raised hopes of opening the black box of traditional networks\cite{yu2023white}. It enables networks to efficiently represent complex multivariate functions by employing the Kolmogorov-Arnold representation theorem\cite{kolmogorov1957representation}. Unlike MLPs, which have fixed activation functions at nodes, KANs use fixed activation functions at edges.
This decomposition helps to reveal the decision-making process and output of the model, which enhances the interpretability of the model. \cite{li2024u} was the first to introduce KANs into visual tasks, reformulating U-Net as U-KAN to improve medical image segmentation and generation.
Despite these initial explorations, the potential of KANs for low-level visual tasks such as low-light image enhancement has not yet been demonstrated.\par
\begin{figure}[t!]\centering
     \includegraphics[height=0.35\textwidth,width=0.48\textwidth]{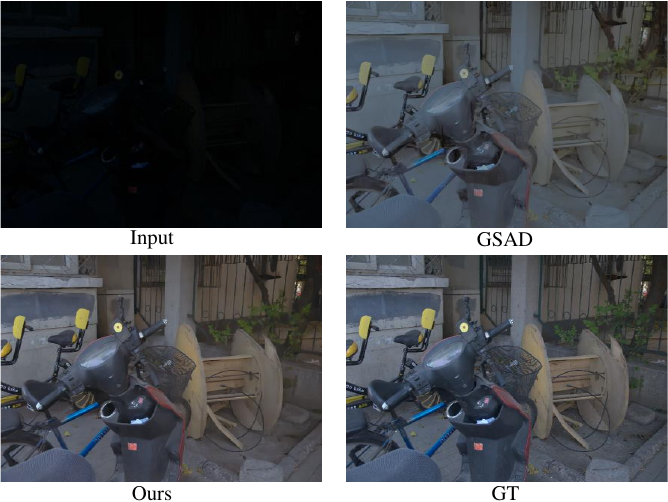}	    
     \caption{Our method effectively learns the nonlinear degradation factors in the low-light domain, especially in darker scenes, and our recovery significantly improves compared to the GSAD.}
    \label{sample}
\vspace{-5mm}
\end{figure}
In this letter, we propose a novel LLIE method (KSID) that introduces KANs to low-level visual tasks for the first time to learn better the nonlinear dependencies between the normal and low-light domains. Specifically, we design the KAN-Block and embed it into the U-Net used for denoising by the diffusion model. KAN-Block consists of the KAN-Layer and DwConv. The KAN-Layer, featuring spline-based convolutional layers and learnable activation functions, effectively captures nonlinear dependencies, significantly enhancing the quality of images generated by the diffusion model. In addition, to improve the stability and visualization of the generation process, we reconstruct the image at each step in the denoising process and introduce frequency domain perception using the Fast Fourier Transform (FFT) to further refine the image details by learning the spectrum of the normal image. As shown in Fig. \ref{sample}, our method shows a significant improvement over the current state-of-the-art method \cite{hou2024global}. We performed extensive experiments on benchmark datasets to demonstrate the effectiveness of our method. Our contribution can be summarized as follows:
\begin{itemize}
\item To the best of our knowledge, we are the first to successfully introduce KANs into the LLIE task, significantly improving the quality of low-light image restoration.

\item We introduce frequency-domain perception for visual orientation enhancement by learning the spectrum of a normal image through the Fast Fourier Transform.
\item We performed extensive experiments on the low-light image enhancement benchmark datasets and achieved impressive performance.
\end{itemize}

\begin{figure*}[ht!]\centering
    \includegraphics[height=0.55\textwidth,width=1\textwidth]{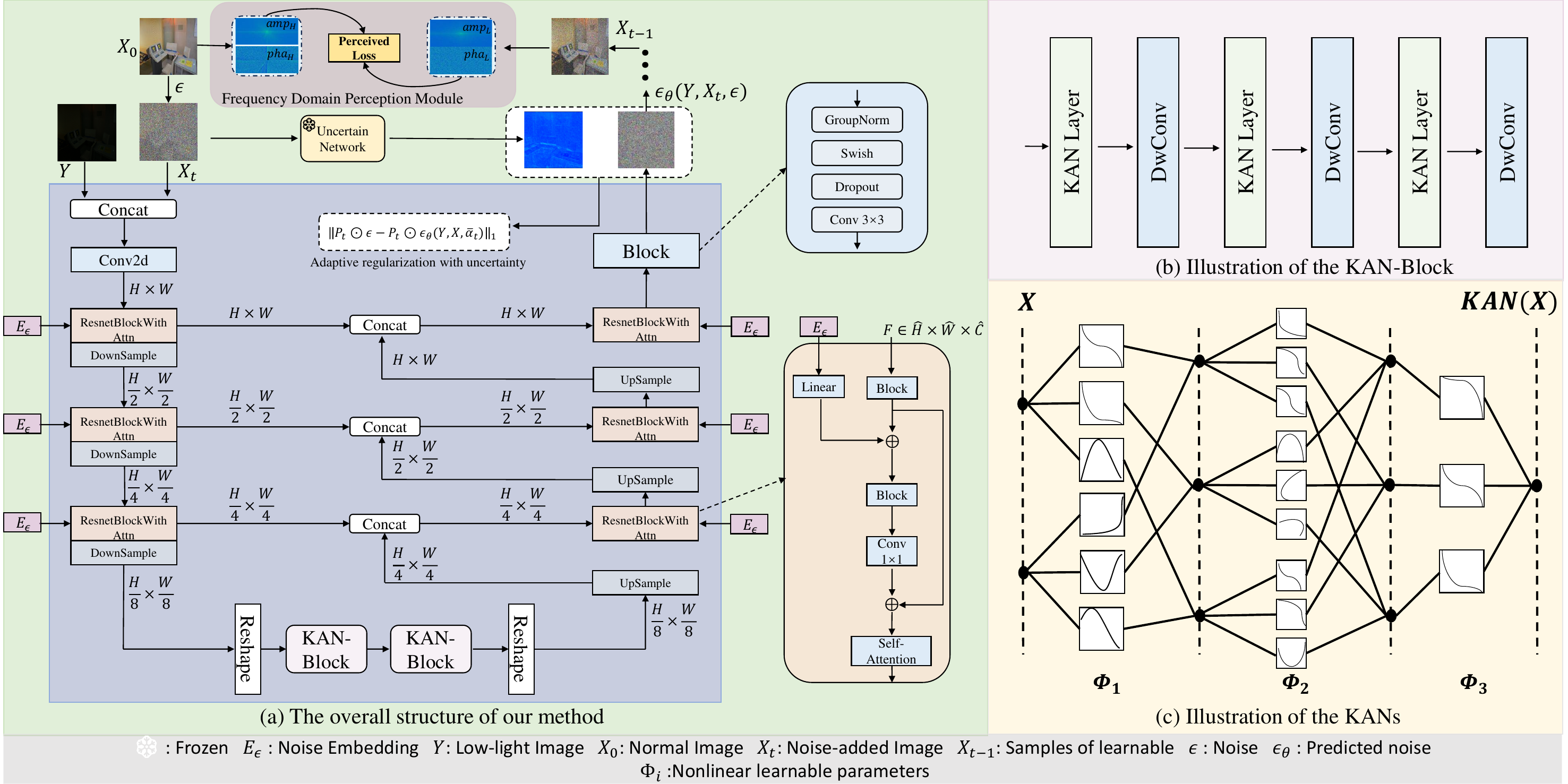}	    
\caption{(a) Illustration of the training workflow of the proposed method. (b) Detailed of the KAN-Block. (c) Structure of the Kolmogorov-Arnold Networks. $X$ is the input feature}
\label{ovall}
\vspace{-5mm}
\end{figure*}
\section{Methods}
\subsection{Kolmogorov-Arnold theorem Preliminaries}
The Kolmogorov-Arnold theorem states that any continuous function can be represented as a composition of a finite number of continuous univariate functions. Specifically, for any continuous function $f(x)$ defined in 
$n-$dimensional real space, where $x=(x_1,x_2,...,x_n)$,  it can be expressed as a composition of a univariate continuous function $h$ and a series of continuous bivariate functions $x_i$ and $g_{q, i}$. Specifically, the theorem shows that there exists such a representation:
\begin{eqnarray}
f(x_1,x_2,...,x_n)=\sum_{q=1}^{2n+1}h(\sum_{i=1}^{n}g_{q,i}(x_i) ) 
\end{eqnarray}
 This representation indicates that even complex functions in high-dimensional spaces can be reconstructed through a series of lower-dimensional function operations.

\subsection{Overall Network Architecture}
The structure of our proposed (KSID) is shown in Fig. \ref{ovall}(a). Our training is divided into two phases: In the first phase, inspired by \cite{hou2024global}, we introduced uncertainty-guided regularization into the diffusion process to enhance the recovery performance in challenging areas; in the second phase, we froze the weights of the uncertainty network to guide the network's learning. In both phases, we utilized the KAN-Block to strengthen the learning of nonlinear dependencies, and in the second phase, we incorporated a frequency-domain perception module to achieve visually-guided enhancement.
\subsection{KAN-Block}
We aim to embed KANs within a low-light image enhancement network to enhance the model’s interpretability and capacity for learning from nonlinear dependencies. As shown in Fig. \ref{ovall}(a), we designed the KAN-Block and integrated it into the U-Net structure for the low-light enhancement task. The U-Net extracts high-level features through a stepwise downsampling operation and recovers low-level details using skip connections. To avoid interference from low-level information, we replace the middle layer in the U-Net with the KAN-Block, with no change in the sampling stage. \par
Specifically, our network begins by taking an input image $X_{0}\in R^{H\times W}$ and adding random noise $\epsilon \in R^{H\times W} $ to obtain the noisy image $X_{t}\in R^{H\times W}$, which is then fed into the U-Net denoising network. Following several downsampling operations and residual concatenations, the image is reshaped and passed into the first KAN-Block. As shown in Fig. \ref{ovall}(b), the KAN-Block consists of DWConv and KAN-Layers. KAN-Block with N KAN-Layers can be represented as:
\begin{eqnarray}
KAN(I)=(\Phi _{N-1}\circ \Phi _{N-2}\circ\cdot \cdot \cdot \circ\Phi _{1}\circ\Phi _{0})I
\end{eqnarray}
where $I$ is the input feature vector; $\Phi_{i}$ signifies the $i$-th KAN-Layer of the entire KAN-Block. In our implementation, the parameter $N$ is set to 3. Unlike the common linear structure, as shown in Fig. \ref{ovall}(c), the weight of each connection in the KAN-Layer is not a simple numerical value but is parameterized as a learnable spline function. Each KAN-Layer $\Phi_{i}$, with $n_{in}$-dimensional input and $n_{out}$-dimensional output, which can be represented as:
\begin{eqnarray}
\Phi=\left \{\phi_{q,p}  \right \}, \verb| |p=1,2,...,n_{in},\verb| | q=1,2,...,n_{out}  
\end{eqnarray}
where $\Phi$ comprises $n_{in} \times n_{out}$ learnable activation functions $\phi$; $\phi_{q,p}$ is the parameter that can be learned. After each KAN-Layer, the features are processed by an efficient depthwise convolutional layer DWConv.
The process of the computation of KAN-Layer from the $i$-th layer to the $i+1$-th layer can be expressed:
 \begin{eqnarray}
I_{i+1} =DwConv(\Phi _{i}(I_{i}))
\end{eqnarray}
The results of the computation can be expressed in the form of a matrix:
\begin{eqnarray}
\Phi _{i}(I_{i})=  
\underbrace{\begin{bmatrix}  
  \phi_{i,1,1(\cdot )}& \phi_{i,1,2(\cdot )}& \cdots  & \phi_{i,1,n_{i}(\cdot )} \\  
  \phi_{i,2,1(\cdot )}& \phi_{i,2,2(\cdot )}& \cdots  & \phi_{i,2,n_{i}(\cdot )} \\  
  \vdots & \vdots & \ddots & \vdots \\  
  \phi_{i,n_{i+1},1(\cdot )}& \phi_{i,n_{i+1},2(\cdot )}& \cdots  & \phi_{i,n_{i+1},n_{i}(\cdot )}  
\end{bmatrix}
}_{\Phi _{i}} I_{i}
\end{eqnarray}
where $\Phi _{i}$ is the function matrix corresponding to the $i$-th KAN-Layer. Features pass through two KAN-Blocks, followed by progressive upsampling to restore the original image size and obtain the predicted noise.
\par

\subsection{Frequency Domain Perception Module}
Although the current low-light enhancement methods based on the diffusion model have made good progress, due to the stochastic nature of their inverse diffusion process and unsatisfactory visual effects, we have introduced frequency-domain perception to make the whole process more stable and achieve visually oriented enhancement.\par
The training of the diffusion denoising probabilistic model starts with obtaining a closed form $X_{t}$ at any time step $t$; in the subsequent process, the model uses a learnable function $\epsilon _{\theta }(Y, X_{t},\bar{\alpha }_{t} )$ to learn the underlying noise distribution\cite{hou2024global}. Since the network can successfully learn this noise, we construct a learnable $X_{t-1}$ for frequency-domain perception, thereby constraining the entire training process to be more stable. It is defined as follows:
\begin{eqnarray}
X_{t-1}=\frac{1}{\sqrt{\alpha _{t}} }(X_{t}-\frac{1-\alpha _{t}}{\sqrt{1-\bar{\alpha } } }\epsilon _{\theta }(Y,X_{t},\bar{\alpha }_{t}) )
\end{eqnarray}
where $\epsilon _{\theta }\in R^{H\times W}$ is the predicted noise distribution. We introduce a frequency domain perceptual loss to learn the spectrum of the normal image.
The Fourier transform can convert an image from the spatial domain to the frequency domain, allowing for better extraction of details and high-frequency information from the image.
The Fourier transform can be defined as follows:
\begin{eqnarray}
& X_{FFT}=\mathcal{F}(X)=A(X)\cdot e^{j\Phi(X)}
\end{eqnarray}
where $A(X)$ represents the magnitude spectrum of the image $X$; $\Phi(X)$ is the phase spectrum; and $\mathcal{F}$ denotes the Fourier transform operation.
Specifically, we first transform the constructed learnable $X_{t-1}$ and normal images from the spatial domain to the frequency domain using the Fast Fourier Transform (FFT). The process is defined as follows:
\begin{eqnarray}
& amp_{high},pha_{high}=\mathcal{F}(X_{0})\\
& amp_{low},pha_{low}=\mathcal{F}(X_{t-1})
\end{eqnarray}
where $amp$ denotes amplitude and $pha$ denotes phase. To align $X_{t-1}$ with $X_{0}$ in high-frequency details, we construct the frequency domain loss $L_{f}$, which is defined as follows:
\begin{eqnarray}
\begin{split}
L_{f}= &\gamma _{1}\left \| amp_{low}-amp_{high} \right \|_{1} + \\
&\gamma _{2}\left \| pha_{low}-pha_{high} \right \|_{1}
\end{split}
\end{eqnarray}
where $\gamma _{1}$ and $\gamma _{2}$ are weighting parameters for amplitude loss and phase loss. By minimizing this loss function, we can make the enhanced image as close as possible to the normal image $X_{0}$ in the frequency domain.
\begin{table*}[]\centering
\caption{Quantitative comparisons of different methods on LOLv1, LOLv2 and LSRW. $\uparrow$ (resp. $\downarrow$) means the larger (resp. smaller), the better.}
\label{table1}
\scalebox{0.78}{
\begin{tabular}{cccccccccccccccc}
\hline
                 &            & \multicolumn{4}{c}{LOLv1}        & \multicolumn{4}{c}{LOLv2-real}   & \multicolumn{4}{c}{LSRW}         &          \\
Methods          & Published  & PSNR$\uparrow$   & SSIM$\uparrow$  & LPIPS$\downarrow$ & FID$\downarrow$     & PSNR$\uparrow$   & SSIM$\uparrow$  & LPIPS$\downarrow$ & FID$\downarrow$     & PSNR$\uparrow$   & SSIM$\uparrow$   & LPIPS$\downarrow$ & FID$\downarrow$    & Param(M) &Times(S)\\ \hline
EnlightenGAN\cite{jiang2021enlightengan}     & TIP’21     & 17.483 & 0.651 & 0.390  & 95.028  & 18.676 & 0.678 & 0.364 & 84.044  & 17.081 & 0.470   & 0.420  & 69.184 & 8.642 &    0.414 \\
RUAS\cite{liu2021retinex}             & CVPR'21    & 16.405 & 0.499 & 0.382 & 102.013 & 15.351 & 0.495 & 0.395 & 94.162  & 14.271 & 0.461  & 0.501 & 78.392 & 0.003 &  - \\
SCI\cite{ma2022toward}              & CVPR'22    & 14.784 & 0.526 & 0.392 & 84.907  & 17.304 & 0.540  & 0.345 & 67.624  & 15.242 & 0.419  & 0.404 & 56.261 & -   & -     \\
URetinex-Net\cite{wu2022uretinex}     & CVPR'22    & 19.842 & 0.824 & 0.237 & 52.383  & 21.093 & 0.858 & 0.208 & 49.836  & 18.271 & 0.518  & 0.419 & 66.871 & -    &-    \\
SNRNet\cite{xu2022snr}           & CVPR'22    & 24.609 & 0.841 & 0.262 & 56.467  & 21.780  & 0.849 & 0.237 & 54.532  & 16.499 & 0.505  & 0.419 & 65.871 & -   &-     \\
Uformer\cite{wang2022uformer}          & CVPR'22    & 19.001 & 0.741 & 0.354 & 109.351 & 18.442 & 0.759 & 0.347 & 98.138  & 16.591 & 0.494  & 0.435 & 82.299 & 20.471  & - \\
Restormer\cite{zamir2022restormer}        & CVPR'22    & 20.614 & 0.797 & 0.288 & 72.998  & 24.910  & 0.851 & 0.264 & 58.649  & 16.303 & 0.453  & 0.427 & 69.219 & -     &-   \\
MIRNet\cite{zamir2022learning}           & TPAMI'22   & 24.140  & 0.842 & \textcolor{red}{0.131} & 69.179  & 21.020  & 0.830  & 0.241 & 49.108  & 16.470  & 0.477  & 0.430  & 93.811 & 31.791    &-  \\
UHDFour\cite{li2023embedding}          & ICLR’23    & 23.093 & 0.821 & 0.259 & 56.912  & 21.785 & 0.854 & 0.292 & 60.837  & 17.300   & 0.529  & 0.433 & 62.032 & -    &-    \\
CLIP-LIT\cite{liang2023iterative}         & ICCV'23    & 12.394 & 0.493 & 0.397 & 108.739 & 15.262 & 0.601 & 0.398 & 100.459 & 13.483 & 0.405  & 0.425 & 77.063 & 0.281     &0.192  \\
NeRCo\cite{yang2023implicit}            & ICCV'23    & 22.946 & 0.785 & 0.311 & 76.727  & 25.172 & 0.785 & 0.338 & 84.534  & \textcolor{red}{19.456} & \textcolor{blue}{0.539}  & 0.423 & 64.555 & 23.385 & 0.756  \\
GSAD\cite{hou2024global}             & NeurIPS'23 & \textcolor{red}{26.402} & \textcolor{blue}{0.875} & \textcolor{blue}{0.188} & \textcolor{red}{40.000}      & \textcolor{blue}{28.805} & \textcolor{blue}{0.894} & \textcolor{blue}{0.201} & \textcolor{blue}{41.456}  & \textcolor{blue}{19.130}  & 0.538  & \textcolor{blue}{0.396} & \textcolor{blue}{57.930}  & 17.435  & 0.486 \\
FourLLIE\cite{wang2023fourllie}             & ACM MM'23 & 24.150 & 0.840 & 0.241 & 58.796      & 22.338 & 0.875 & 0.233 & 45.821  & \textcolor{red}{19.870}  & \textcolor{red}{0.602}  & 0.437 & 70.255  & 0.120  & 0.031 \\
Lightendiffusion\cite{jiang2024lightendiffusion} & ECCV'24    & 20.188 & 0.814 & 0.316 & 85.930   & 22.697 & 0.853 & 0.306 & 75.582  & 18.397 & 0.534  & 0.428 & 67.801 & 27.835  &  0.568\\
UPT-Flow\cite{xu2024upt}         & PR'24      & 20.644 & 0.865 & 0.215 & 48.926  & 25.056 & 0.889 & 0.231 & 20.757  & -      & -      & -     & -      & 23.436  & 1.011 \\
Ours             &            & \textcolor{blue}{26.161} & \textcolor{red}{0.877} & 0.192 & \textcolor{blue}{40.890}   & \textcolor{red}{31.154} & \textcolor{red}{0.934}  & \textcolor{red}{0.175} & \textcolor{red}{29.259}  & 18.650  & 0.547 & \textcolor{red}{0.393} & \textcolor{red}{55.860}  & 21.779  &  0.526\\ \hline
\end{tabular}}
\end{table*}

\begin{figure*}[ht!]\centering
     \includegraphics[height=0.22\textwidth,width=1\textwidth]{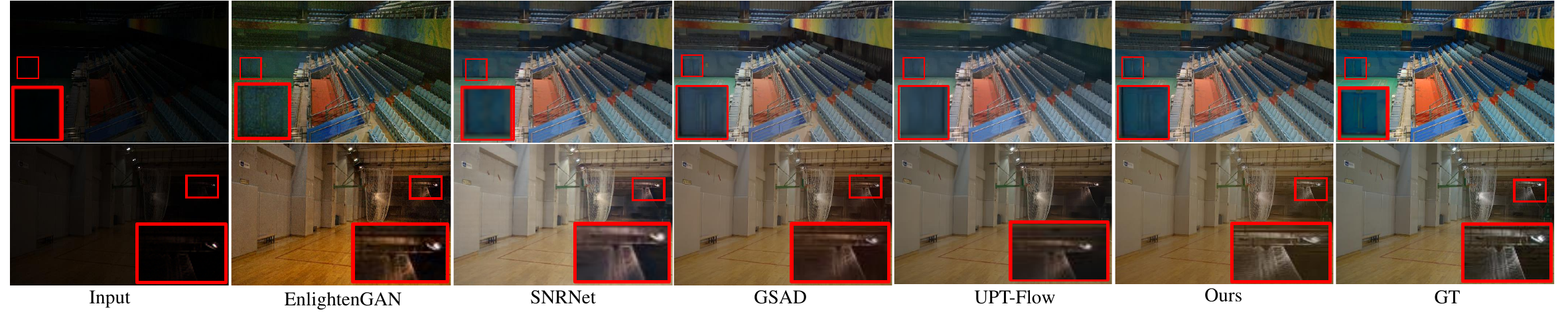}	    
     \caption{Visual comparisons of the enhanced results by different methods on LOLv2. }
    \label{vis}

\end{figure*}
\section{Experiments}
\subsection{Experimental Settings}
\subsubsection{Datasets and Metrics}
We used three common low-light image enhancement benchmark datasets for evaluation: LOLv1, LOLv2, and LSRW. For evaluation metrics, we use the Peak Signal-to-Noise Ratio (PSNR) and Structural Similarity (SSIM) as two full-reference distortion metrics.
In addition, we use Learned Perceptual Image Block Similarity (LPIPS) and Fréchet Inception Distance (FID) as two perceptual metrics.
We used model parameter count Param(M) and average inference time Times(S) to evaluate the efficiency of the model.

\subsubsection{Implementation Details}
We implemented KSID on an NVIDIA RTX 3090 GPU with PyTorch, setting the batch size to 8 and patch size to 96$\times$96. The learning rate was set to 1e-4, with Adam as the optimizer. The training process is divided into two phases: the first phase focuses on optimising the uncertainty network, with the number of epochs set to 1e6; the second phase extends the training by setting the number of epochs to 2e6.
\subsection{Comparisons With State-of-the-Art Methods}
We qualitatively and quantitatively compare the proposed KSID with state-of-the-art low-light image enhancement methods. We train on the LOLv1 dataset and test on all datasets.
\subsubsection{Quantitative Comparisons}
Table \ref{table1} presents the quantitative results of various LLIE methods, indicating that our approach is competitive in the metrics PSNR, SSIM, LPIPS and FID. Notably, on the LOLv2-real dataset, our method achieves state-of-the-art performance in both SSIM and PSNR (full-reference metrics) as well as FID and LPIPS (perceptual metrics).
Additionally, our method achieves the best results on the large-scale LSRW dataset in SSIM, LPIPS, and FID metrics, further confirming its strong generalization and robustness.
\subsubsection{Qualitative Comparisons}
We performed a qualitative comparison with different LLIE methods.
As shown in Fig. \ref{vis}, in the LOLV2-real test dataset, we observed that the images restored by other methods suffered from colour distortion and could not effectively handle uneven illumination. However, our method has effectively addressed these issues with restored images closer to the reference image color distribution and better at recovering detailed information.

\begin{figure}[ht!]\centering
     \includegraphics[height=0.14\textwidth,width=0.4\textwidth]{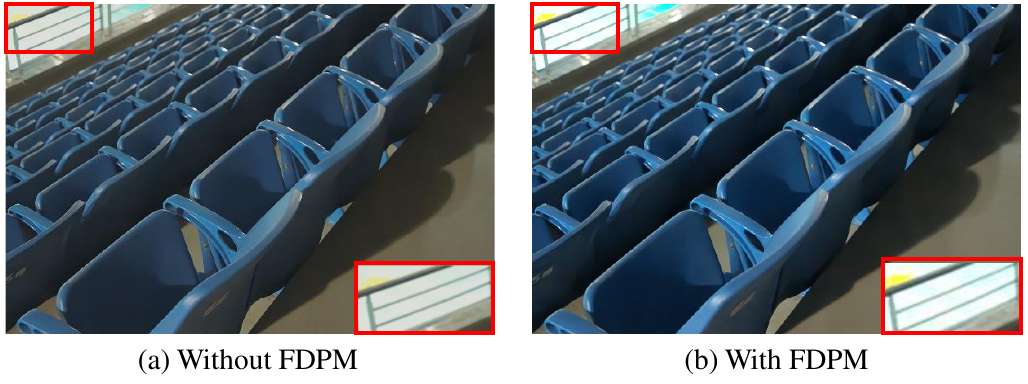}	    
     \caption{A visual comparison of results with and without the Frequency Domain Perception Module.}
    \label{fre}
    \vspace{-3mm}

\end{figure}

\begin{table}[]\centering

\caption{Ablation experiments were conducted for different modules of KSID. $\uparrow$ (resp. $\downarrow$) means the larger(resp. smaller), the better.}
\label{ablation}
\scalebox{0.82}{
\begin{tabular}{ccccccc}
\hline
KAN-Block & FDPM & PSNR$\uparrow$   & SSIM$\uparrow$  & LPIPS$\downarrow$ & FID$\downarrow$   & Param(M) \\ \hline
$\times$         & $\times$    & 28.805 & 0.894 & 0.201 & 41.456  &17.435 \\
$\checkmark$         & $\times$    & \textcolor{red}{31.915} & 0.930 & 0.189 & 32.640 &21.779 \\
$\times$         & $\checkmark$    & 29.105 & 0.897 & 0.179  & 30.575 & 17.435 \\
$\checkmark$         & $\checkmark$    & 31.154 & \textcolor{red}{0.934} & \textcolor{red}{0.175} & \textcolor{red}{29.259} & 21.779 \\ \hline
\end{tabular}
}
\vspace{-5mm}
\end{table}
\subsection{Ablation Study}
To evaluate the effectiveness of our model for low-light image enhancement tasks, we performed ablation studies on different modules on the LOLv2 test set. Table \ref{ablation} demonstrates that the KAN-Block significantly enhances the model's ability to learn the nonlinear degradation relationship between low-light and normal images, resulting in restored images that closely align with the true distribution. As shown in Fig. \ref{fre}, we present a comparison of results with and without the FDPM. It is evident that our FDPM significantly enhances the visual quality of the improved images.

\section{Conclusion}
In this paper, we propose a novel low-light image enhancement method, KSID, which introduces KANs into the LLIE task for the first time, improves the model's ability to learn nonlinear dependencies and achieves high-quality mapping of the degradation parameters. In addition, we introduce the Frequency Domain Perception Module to refine the image details further and make the inverse diffusion process more stable. Extensive experiments validate the effectiveness and robustness of our method. Overall, we provide an initial exploration of the potential of KANs in the field of LLIE and argue that this non-traditional linear network structure is important for processing low-level visual tasks.


\ifCLASSOPTIONcaptionsoff
  \newpage
\fi

\vfill
\pagebreak

\bibliographystyle{plain}
\bibliography{template}

\end{document}